\documentclass[10pt,twocolumn,letterpaper]{article}

\usepackage[
    top=0.75in,
    bottom=0.75in,
    left=0.7in,
    right=0.7in
]{geometry}

\usepackage[hyphens]{url}
\usepackage{graphicx}
\usepackage{amsmath,amssymb}
\usepackage{booktabs}
\usepackage{multirow}
\usepackage{pifont}
\usepackage{xcolor}
\usepackage{microtype}
\usepackage{natbib}
\usepackage[hidelinks]{hyperref}
\usepackage{authblk}
\usepackage{orcidlink}

\newcommand{\method}{VTM-Nav}

\title{
\method: Harnessing Cross-Episode Experience for Object-Goal Navigation
with Hierarchical Visual-Topological Memory
}

\author[1,2]{Xiaoran Xu$^*$\,\orcidlink{0009-0001-8350-2828}}
\author[1,2]{Yupeng Wu$^*$\,\orcidlink{0000-0001-5453-9755}}
\author[3]{Tianyu Xue\,\orcidlink{0009-0008-2610-826X}}
\author[1,2]{Yifan Xu\,\orcidlink{0000-0003-2467-888X}}
\author[2]{Xuanran Dong\,\orcidlink{0009-0000-4503-6320}}
\author[1,2]{Xiaoshan Yang$^\dagger$\,\orcidlink{0000-0001-5453-9755}}
\author[1,2]{Changsheng Xu\,\orcidlink{0000-0001-8343-9665}}

\affil[1]{MAIS, Institute of Automation, Chinese Academy of Sciences, Beijing, China}
\affil[2]{School of Advanced Interdisciplinary Sciences, University of Chinese Academy of Sciences, Beijing, China}
\affil[3]{Tsinghua University, Beijing, China}

\affil[ ]{
\texttt{xuxiaoran22@mails.ucas.ac.cn},
\texttt{wuyupeng23@mails.ucas.cn},
\texttt{xuety23@mails.tsinghua.edu.cn},
\texttt{dongxuanran24@mails.ucas.ac.cn}\\
\texttt{\{yifan.xu, xiaoshan.yang, changsheng.xu\}@nlpr.ia.ac.cn}
}

\affil[ ]{\small $^*$Equal contribution. \quad
$^\dagger$Corresponding author.}

\date{}

\begin{document}

\maketitle

\begin{abstract}
Training-free ObjectNav agents increasingly use vision-language models (VLMs), yet typically discard acquired scene knowledge after each request. We study cross-episode ObjectNav, where each request is an independently initialized, single-goal episode and only self-acquired, scene-scoped memory persists across episodes. We ask whether an agent with fixed model parameters and navigation components can reuse such experience without retraining or oracle information.
We introduce \method, a training-free framework with a persistent Hierarchical Visual-Topological Memory (VTM). VTM uses a coarse room topology to index room-owned visual memories, distinguishes in-room from remote-visible evidence, and retains successful approach cues. For each request, \method{} re-localizes the agent in accumulated scene structure, retrieves target-relevant records from plausible rooms, and grounds memory guidance in candidates derived from the current observation. A conservative execution guard further handles local failures.
Under matched 40-step comparisons, \method{} exceeds the memory-reset WMNav control by 4.6, 2.0, and 0.8 SR points on HM3D v0.1, HM3D v0.2, and MP3D, respectively, with comparable or higher SPL. On HM3D, it also exceeds WMNav harnessed by textual memory by 3.1 and 5.5 SR points. These results demonstrate effective reuse of cross-episode scene experience through hierarchical visual-topological memory.
\end{abstract}

\section{Introduction}
Object-goal navigation (ObjectNav) requires an embodied agent to navigate in an indoor scene, find an instance of a target object category, and stop near it \cite{anderson2018evaluation,batra2020objectnav}. Recent training-free agents use vision-language models (VLMs) to exploit open-vocabulary semantics and object-room priors without task-specific training \cite{vlmnav,vlfm,wmnav,mernav}.

\begin{figure}[t]
  \centering
  \includegraphics[width=0.48\textwidth]{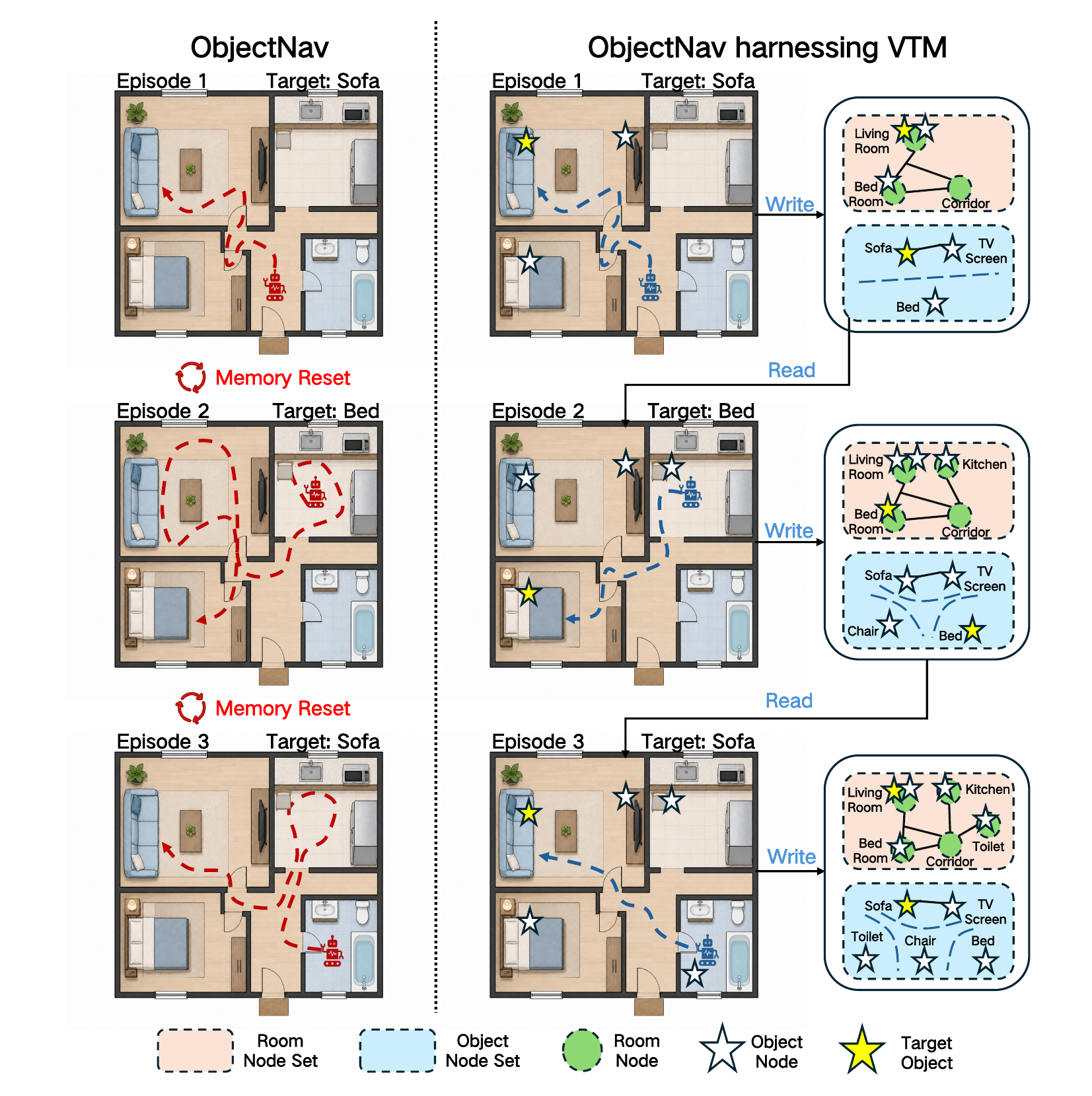}
    \caption{Harnessing cross-episode experience for ObjectNav in a repeatedly visited scene. Conventional memory-reset agents discard scene knowledge after each request, whereas \method{} retains a room topology and visual experience to guide later, independently initialized requests.}
  \label{fig:motivation}
\end{figure}

Despite this progress, conventional ObjectNav evaluation resets scene-specific state after every run: each request starts from a random pose, specifies one target category, and discards acquired knowledge at termination \cite{anderson2018evaluation,batra2020objectnav}. This convention supports controlled benchmarking but underuses experience available in recurring deployment. A service robot revisiting the same home, office, or hotel could reuse discovered room connectivity, object viewpoints, remote visual cues, and successful approaches instead of treating every request as new.

Figure~\ref{fig:motivation} illustrates our \emph{cross-episode ObjectNav} setting. Each request remains an independently initialized, single-goal ObjectNav episode; its observation and action spaces, success criterion, and target vocabulary are unchanged. At an episode boundary, physical navigation state and short-term execution history are reset, and the only state allowed to persist is scene-scoped experience acquired from the agent's observations, actions, and outcomes. We use \emph{harnessing cross-episode experience} to mean more than retaining this state: the agent must re-localize in previously acquired scene structure, retrieve visual evidence, ground that evidence in the current observation, and use it to guide a later episode without parameter updates.

Most memory-augmented methods build working, global-to-ego, or topological representations for the current navigation process \cite{memonav,mem2ego,dynavlm,etpnav,wmnav}, while persistent-navigation work commonly studies continuous goal sequences \cite{goatbench,ssmgnav}. Cross-episode reuse removes that physical-state continuity. It therefore requires a memory that can support both re-localization after independent initialization and target-conditioned retrieval, without ground-truth object locations, oracle semantic maps or room labels, shortest paths, or parameter updates.

We present Visual-Topological Memory(VTM). VTM uses a coarse room topology as a structural index over fine-grained, room-owned visual memories. The topology captures observed connectivity and supports re-localization and coarse retrieval; each room memory preserves object evidence, representative viewpoints, remote-visible cues, and successful approaches for fine-grained retrieval.

We propose \method, a training-free VLM framework that harnesses cross-episode experience through this hierarchy. Retrieval first identifies plausible rooms from the stored topology and then queries the corresponding room-owned visual memory. Retrieved experience enters the VLM prompt as soft context and can re-rank only candidates derived from the current observation. A conservative execution guard reviews no-progress behavior, oscillations, blocked approaches, and premature stopping while preserving \texttt{stop} when the target is visible.

We evaluate \method{} under a controlled cross-episode protocol on HM3D v0.1, HM3D v0.2, and MP3D. Our contributions are:
\begin{itemize}
\item We formulate cross-episode ObjectNav as experience reuse across independently initialized, single-goal episodes, where only self-acquired, scene-scoped memory persists.
\item We introduce a Hierarchical Visual-Topological Memory that couples a coarse room topology with fine-grained, room-owned visual memories, preserving the spatial provenance of object evidence and successful approaches.
\item We harness this experience through re-localization, target-conditioned coarse-to-fine retrieval, observation-grounded candidate re-ranking, and a conservative execution guard, without updating the VLM or navigation components.
\item Under controlled 40-step comparisons with matched navigation components, we evaluate hierarchical VTM against a memory-reset control on three ObjectNav benchmarks and against persistent textual memory on the two HM3D benchmarks.
\end{itemize}

\section{Related Work}

\subsection{Training-Free Object-Goal Navigation with VLMs}

Earlier ObjectNav systems relied on learned policies, semantic exploration, task-specific supervision, or offline representation learning \cite{habitatweb,ovrl,semantic_mapping}. Recent zero-shot or training-free methods instead use open-vocabulary perception, foundation models, and VLM reasoning \cite{cows,vlfm,vlmnav,wmnav,mernav,clcotnav,goalvlm}. Although these models provide strong semantic priors, most evaluations discard connectivity, viewpoints, and approach traces after each episode. We keep the VLM backbone and action pipeline fixed to test whether self-acquired scene experience improves cross-episode navigation in the same scene at comparable path efficiency.



\begin{figure*}[ht]
  \centering
  \includegraphics[width=0.98\textwidth]{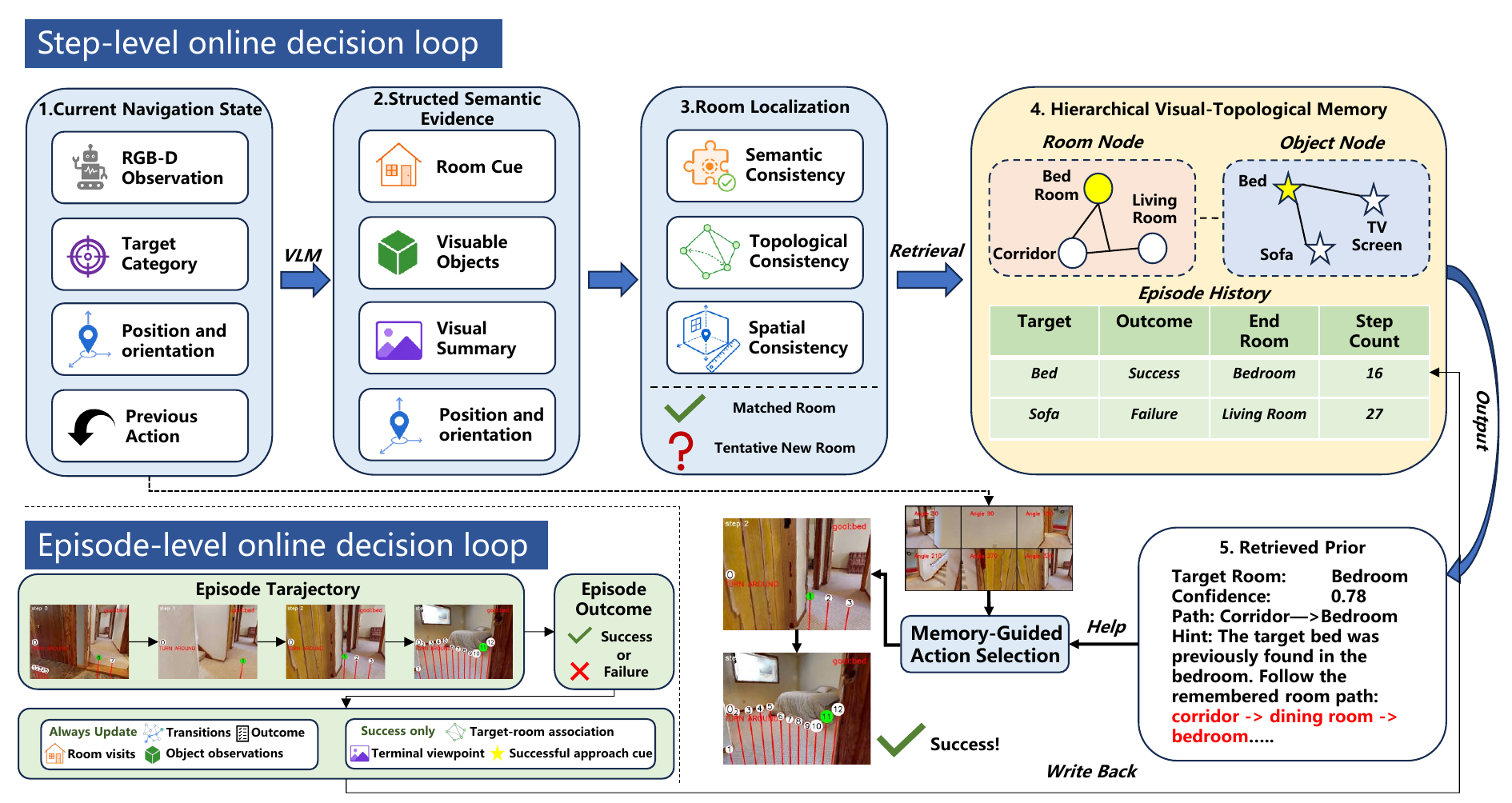}
  \caption{
Overview of \method{}. The upper branch shows re-localization, Hierarchical Visual-Topological Memory update, target-conditioned retrieval, and observation-grounded action selection. The lower branch shows how episode outcomes update room topology, visual object records, and success-backed approach cues for later independently initialized episodes.
}
\label{fig:method}
\end{figure*}


\subsection{Topological and Structured Memory for Navigation}

Topological and scene-graph methods organize places, connectivity, objects, and semantic relations, while visual memories retain observations or goal-relevant features \cite{topological_navigation,etpnav,sgnav,openscenegraphs,esceme,memonav,mem2ego}. Recent VLM-based methods further model executable relations or predicted world states \cite{dynavlm,wmnav}.

VTM couples these representations in a two-level hierarchy. A coarse room graph captures connectivity and indexes room-owned visual memories containing object evidence, viewpoints, summaries, and success-backed approaches. The records preserve in-room or remote-visible provenance. After episode reset, the agent re-localizes in the graph, retrieves a plausible target room, and queries its local visual memory.

\subsection{Persistent and Cross-Episode Navigation}

GOAT-Bench and SSMG-Nav reuse experience along continuous goal sequences, with each subtask inheriting the preceding terminal state and spatial reference \cite{goatbench,ssmgnav}. Continual ObjectNav instead studies adaptation to goal categories introduced over stages \cite{cnav,continual_learning,continual_robot_learning}.

In contrast, cross-episode ObjectNav uses independently initialized, single-goal episodes while retaining only scene-scoped memory. The target vocabulary, model parameters, perception stack, and controller remain fixed. The agent must therefore re-localize in the stored room graph before retrieving target-relevant, room-owned visual evidence.

\section{Method}

\subsection{Problem Formulation}
We study \emph{cross-episode experience reuse} while preserving standard ObjectNav within every episode. The observation and action spaces, success criterion, target vocabulary, model parameters, and navigation components remain fixed. Episodes are independently initialized, and physical navigation state, current observations, and short-term execution history do not cross their boundaries. The only persistent state is self-acquired, scene-scoped memory.

In each episode, the agent starts at position $P_0$ in environment $\mathcal{E}$ with target $g\in\mathcal{G}$. At step $t$, it observes $O_t$, estimates position $P_t$, and receives $K_t$ navigable candidates $\mathcal{C}_t=\{c_{t,k}\}_{k=1}^{K_t}$. Each candidate has an agent-centric polar displacement $(r_{t,k},\theta_{t,k})$ and textual description $d_{t,k}$. The agent traverses a candidate or issues \texttt{stop}, succeeding when it stops within the success-distance threshold $d_{\mathrm{succ}}$ of target.

For scene $s$, the cross-episode sequence $E_s=(e_{s,1},\ldots,e_{s,J_s})$ contains $J_s$ episodes with episode-specific start positions $P_{s,j,0}$ and targets $g_{s,j}$. Scene memory $\mathcal{M}_s$ is initialized as empty and updated after each episode:
\begin{equation}
\mathcal{M}_s^{(j)}
=
\Phi_{\mathrm{write}}
\left(
\mathcal{M}_s^{(j-1)},
\tau_{s,j}
\right),
\qquad
\mathcal{M}_s^{(0)}=\varnothing,
\end{equation}
where $\Phi_{\mathrm{write}}$ is the memory-write operator and $\tau_{s,j}$ contains the episode's egocentric observations, estimated positions and orientations, semantic predictions, executed actions, and outcome. Memory is restricted to navigation-time information and excludes ground-truth object locations, oracle semantic maps or room labels, and shortest paths. It is scene-scoped and cannot be queried across scenes. Harnessing cross-episode experience means that a later episode may consult $\mathcal{M}_s^{(j-1)}$ after re-localization, while its per-episode task definition and success criterion remain unchanged.

\subsection{Overview of \method}

\method{} adds a scene-scoped experience layer to the training-free WMNav backbone, which constructs depth-based navigable candidates and uses a VLM for action selection \cite{wmnav,vlmnav}. As shown in Figure~\ref{fig:method}, \method{} harnesses experience through a re-localize -> update -> retrieve -> ground -> act -> update loop: current semantic evidence re-localizes the agent and updates room-owned records before target-conditioned retrieval guides candidates grounded in the current observation. Since retrieval follows the current-step update, it combines cross-episode memory with current evidence.

Let $\mathcal{M}_{s,j,t}$ denote the scene memory before step $t$ of episode $e_{s,j}$. The online flow is
\begin{equation}
z_t \rightarrow \lambda_t \rightarrow \mathcal{M}_{s,j,t}^{+}
\rightarrow \rho_t \rightarrow \hat{a}_t
\rightarrow \tilde{a}_t \rightarrow a_t,
\end{equation}
where $z_t$ is semantic evidence, $\lambda_t$ is room localization, and $\mathcal{M}_{s,j,t}^{+}$ is the post-write memory. Retrieval produces prior $\rho_t$; the VLM proposes $\hat{a}_t$, memory guidance yields $\tilde{a}_t$, and the execution guard produces $a_t$ using short-term history $h_t$.

At termination, the trajectory and outcome update the persistent scene memory for the next independently initialized episode; the VLM and controller remain fixed, and short-term history is reset.

\subsection{Hierarchical Visual-Topological Memory}

\textbf{Semantic Evidence.}
At each step, Qwen3-VL-Plus processes the current egocentric RGB observation using a fixed structured prompt and returns a room label, a set of visible object categories, and a compact natural-language scene summary. We refer to these observation-grounded outputs as visual-semantic records. 
The semantic perception shown as:
\begin{equation}
z_t =
(\hat{r}_t,\hat{\mathcal{Q}}_t,u_t,P_t,R_t,a_{t-1},g),
\end{equation}
where $\hat{r}_t$ is the perceived room category, $\hat{\mathcal{Q}}_t$ is the normalized visible-object set, and $u_t$ is a textual visual-semantic summary generated from the current RGB observation. $P_t$ and $R_t$ are the estimated position and orientation, $a_{t-1}$ is the previous action, and $g$ is the target category. Before writing, we remove room-category mentions misclassified as objects, discard negated object mentions, and keep ObjectNav targets in the object vocabulary. Objects seen through openings are marked as remote-visible rather than in-room evidence.\\
\textbf{Hierarchical Memory Structure.} 
VTM uses a two-level room-to-object hierarchy. Its coarse-level room topology is
\begin{equation}
\mathcal{T}_s =
(\mathcal{V}^{\mathrm{room}}_s,\mathcal{E}^{\mathrm{room}}_s),
\label{eq:room_topology}
\end{equation}
where $\mathcal{V}^{\mathrm{room}}_s$ contains room nodes and $\mathcal{E}^{\mathrm{room}}_s$ contains observed transitions. Each room node stores a canonical room label, semantic aliases, localization confidence, visit statistics, and recent visual summaries.

Each room $v\in\mathcal{V}^{\mathrm{room}}_s$ owns a local object memory
\begin{equation}
\mathcal{O}_{s,v} =
(\mathcal{O}^{\mathrm{in}}_{s,v},
\mathcal{O}^{\mathrm{rem}}_{s,v}),
\label{eq:local_object_memory}
\end{equation}
where $\mathcal{O}^{\mathrm{in}}_{s,v}$ contains directly observed in-room objects and $\mathcal{O}^{\mathrm{rem}}_{s,v}$ contains cues observed through openings.

The scene-level object memory is
\begin{equation}
\mathbb{O}_s =
\{\mathcal{O}_{s,v}
\mid
v\in\mathcal{V}^{\mathrm{room}}_s\}.
\label{eq:scene_object_memory}
\end{equation}
The complete scene memory is
\begin{equation}
\mathcal{M}_s =
(\mathcal{T}_s,\mathbb{O}_s,\mathcal{H}_s),
\label{eq:scene_memory}
\end{equation}
where $\mathcal{H}_s$ stores episode-level search outcomes and navigation history.
Each object node stores its category, confidence, observation count, viewpoints, visual summaries, navigation hints, and success-backed approaches. Remote-visible nodes remain associated with the observer room and optionally record an opening relation and destination-room hypothesis. This room ownership supports coarse-to-fine retrieval from a target-bearing room to its local object memory.\\
\textbf{Room Localization.} Given evidence $z_t$, VTM scores each candidate room $v\in\mathcal{V}^{\mathrm{room}}_s$:
\begin{equation}
S_{\mathrm{loc}}(v,z_t)
=
S_{\mathrm{sem}}(v,z_t)
+
S_{\mathrm{topo}}(v,z_t)
+
S_{\mathrm{sp}}(v,z_t),
\end{equation}
where the three terms measure semantic, topological, and spatial consistency, respectively. Each lies in $[0,1]$, and we use their unweighted sum, so $S_{\mathrm{loc}}\in[0,3]$; this fixed rule is shared across datasets.

Semantic consistency averages room-name, object-overlap, and visual-summary scores:
\begin{equation}
S_{\mathrm{sem}}(v,z_t)
=
\frac{1}{3}
\left(
S_{\mathrm{name}}(v,z_t)
+
S_{\mathrm{ovl}}(v,z_t)
+
S_{\mathrm{desc}}(v,z_t)
\right).
\end{equation}
Let $\mathcal{A}_v$, $\mathcal{C}_v$, and $\mathcal{U}_v$ denote the canonical room label with its aliases, stored object-category set, and recent visual-summary set, respectively:
\begin{equation}
\begin{aligned}
S_{\mathrm{name}}(v,z_t)
&=\mathbb{I}[\hat{r}_t\in\mathcal{A}_v],\\
S_{\mathrm{ovl}}(v,z_t)
&=J(\hat{\mathcal{Q}}_t,\mathcal{C}_v),\\
S_{\mathrm{desc}}(v,z_t)
&=\max_{u_i\in\mathcal{U}_v}J(\mathrm{kw}(u_t),\mathrm{kw}(u_i)).
\end{aligned}
\end{equation}
Here, $\mathbb{I}$ is the indicator function, $J(A,B)=|A\cap B|/|A\cup B|$ is set to zero when $A\cup B=\varnothing$, and $\mathrm{kw}$ extracts normalized keywords. $S_{\mathrm{desc}}$ is zero when $\mathcal{U}_v=\varnothing$. Confirmed in-room objects populate $\mathcal{C}_v$ before remote-visible fallback cues.

Topological consistency uses the previously localized room $v_{t-1}^\star$:
\begin{equation}
S_{\mathrm{topo}}(v,z_t)
=
\mathbb{I}\left[v=v_{t-1}^\star\ \lor\ (v_{t-1}^\star,v)\in\mathcal{E}^{\mathrm{room}}_s\right],
\end{equation}
If no previous room localization is available, $S_{\mathrm{topo}}$ is set to zero. Spatial consistency compares position $P_t$ with recent positions assigned to $v$:
\begin{equation}
S_{\mathrm{sp}}(v,z_t)
=
\max_{P_i\in\mathcal{P}_v}
\frac{1}{1+\lVert P_t-P_i\rVert_2},
\end{equation}
where $\mathcal{P}_v$ is the set of stored positions for room $v$; $S_{\mathrm{sp}}$ is zero when $\mathcal{P}_v=\varnothing$.

The best-matching room is
\begin{equation}
v_t^\star
=
\arg\max_{v\in\mathcal{V}^{\mathrm{room}}_s}
S_{\mathrm{loc}}(v,z_t).
\end{equation}
The argmax is evaluated when $\mathcal{V}^{\mathrm{room}}_s\neq\varnothing$; otherwise, the evidence remains tentative. A sufficiently high score assigns $v_t^\star$. Repeated tentative support or one high-confidence observation promotes a new room, preventing transient VLM descriptions from polluting the topology. This localization result, including a tentative outcome, is denoted by $\lambda_t$.\\
\textbf{Object-Node Association.} Once $\lambda_t$ assigns an existing or newly promoted room $v_t^\star$, each observation $q\in\hat{\mathcal{Q}}_t$ is classified as
\begin{equation}
\kappa_t(q)
\in
\{\mathrm{in},\mathrm{rem}\},
\label{eq:evidence_type}
\end{equation}
and written to the corresponding local memory of $v_t^\star$. Its best-matching node is
\begin{equation}
o_t^\star(q)
:=
\arg\max_{o\in
\mathcal{O}^{\kappa_t(q)}_{s,v_t^\star}}
S_{\mathrm{obj}}(o,q),
\label{eq:object_match}
\end{equation}
The argmax is evaluated only when $\mathcal{O}^{\kappa_t(q)}_{s,v_t^\star}\neq\varnothing$; an empty local set creates a new node directly. Here, $S_{\mathrm{obj}}$ measures category agreement and visual-description compatibility, and $\tau_{\mathrm{obj}}$ is the fixed association threshold. A score above $\tau_{\mathrm{obj}}$ updates the matched node; otherwise, a new node is created under $v_t^\star$.
A remote-visible node remains attached to the observer room and records the opening cue and, when inferable, a hypothesized destination room.\\
\textbf{Memory Update.} After localization, VTM updates the matched room's visit count, confidence, and visual summaries. Promoted tentative evidence creates a new room node:
\begin{equation} 
\mathcal{V}^{\mathrm{room}}_s \leftarrow \mathcal{V}^{\mathrm{room}}_s \cup \{v_{\mathrm{new}}\}. 
\end{equation}
A newly confirmed room transition inserts the corresponding edge:
\begin{equation} 
\mathcal{E}^{\mathrm{room}}_s \leftarrow \mathcal{E}^{\mathrm{room}}_s \cup \{e_{\mathrm{new}}\}. 
\end{equation}
For transition edge $(v_i,v_j)$, let $\gamma_{ij}$ be its confidence, $n_{ij}$ its observation count, and $\Delta_e>0$ the fixed confidence increment. They update as
\begin{equation}
\gamma_{ij}
\leftarrow
\min(1,\gamma_{ij}+\Delta_e),
\qquad
n_{ij}
\leftarrow
n_{ij}+1.
\end{equation}

For object node $o$, let $\gamma_o$ be its confidence and $\Delta_o^{0}>0$ the fixed increment for a direct in-room observation. The update is
\begin{equation}
\gamma_o
\leftarrow
\min(1,\gamma_o+\Delta_o^{0}).
\end{equation}
After a successful episode, the final target entry becomes success-backed and receives the additional confidence increment $\Delta_o^{+}$:
\begin{equation}
\gamma_o
\leftarrow
\min(1,\gamma_o+\Delta_o^{+}),
\qquad
\Delta_o^{+}>\Delta_o^{0}>0.
\end{equation}
Its viewpoint, visual summary, and approach direction are also stored. The resulting records define the post-write memory $\mathcal{M}_{s,j,t}^{+}$.

\subsection{Memory Retrieval and Action Guidance}
\textbf{Hierarchical Retrieval.}
Given target category $g$, VTM first searches the room-owned object records in the post-write scene memory. For room $v$, confirmed in-room evidence is ranked by
\begin{equation}
R_{\mathrm{in}}(v,g)
=
\gamma_{v,g}
+\eta_1 n_{v,g}
+\eta_2 n^{\mathrm{succ}}_{v,g},
\end{equation}
where $\gamma_{v,g}$ is the stored confidence, $n_{v,g}$ is the observation count, and $n^{\mathrm{succ}}_{v,g}$ is the number of success-backed records. If no confirmed in-room record is available, VTM falls back to remote-visible evidence, whose score uses its confidence and observation count with a lower confidence cap. This ordering prevents an object glimpsed through an opening from overriding evidence confirmed inside a room.

Let $v_t$ be the localized current room and $v_g$ the retrieved target-bearing room. When they are connected in the stored room topology, VTM computes
\begin{equation}
\pi_t
=
\operatorname{BFS}
\left(
\mathcal{T}_s,v_t,v_g
\right),
\end{equation}
which gives the shortest observed room-hop path. If the current room is not reliably localized or no connection has been observed, $\pi_t$ remains empty. The retrieved prior $\rho_t$ contains the evidence type, target room, confidence, available visual-semantic viewpoint record, and optional room path.\\
\textbf{Observation-Grounded Action Guidance.}
The retrieved prior is provided to the VLM as soft context and can also re-rank the navigable candidates derived from the current observation. Let $\hat{k}_t$ be the candidate selected by the original VLM, and let
\begin{equation}
k_t^\dagger
=
\arg\max_{k\in\{1,\ldots,K_t\}}
B(d_{t,k},\rho_t),
\end{equation}
where $B$ measures lexical agreement between candidate description $d_{t,k}$ and the remembered room, object, and navigation cues, while penalizing blocked or non-navigable structures. The memory-guided selection is
\begin{equation}
\tilde{k}_t
=
\begin{cases}
k_t^\dagger,
&
B(d_{t,k_t^\dagger},\rho_t)\geq\tau_{\mathrm{bias}},\\
\hat{k}_t,
&
\text{otherwise}.
\end{cases}
\end{equation}

A conservative execution guard handles local failures not addressed by long-term memory, including blocked motion, no progress, revisits, oscillation, and premature stopping. It remains inactive by default and preserves \texttt{stop} when the target is visible. The lexical dictionaries, confidence caps, thresholds, tie-breaking rules, and complete guard conditions are provided in the Appendix.

\begin{table*}[t]
\centering
\small
\setlength{\tabcolsep}{2.5pt}
\begin{tabular}{llccccccc}
\toprule
\multirow{2}{*}{Method} & \multirow{2}{*}{Foundation model} & \multirow{2}{*}{Steps} &
\multicolumn{2}{c}{HM3D v0.1} & \multicolumn{2}{c}{HM3D v0.2} & \multicolumn{2}{c}{MP3D} \\
\cmidrule(lr){4-5}\cmidrule(lr){6-7}\cmidrule(lr){8-9}
& & & SR $\uparrow$ & SPL $\uparrow$ & SR $\uparrow$ & SPL $\uparrow$ & SR $\uparrow$ & SPL $\uparrow$ \\
\midrule
\multicolumn{9}{l}{\textit{Task-trained and non-zero-shot methods}} \\
Habitat-Web \cite{habitatweb} & -- & 500 & 41.5 & 16.0 & -- & -- & 31.6 & 8.5 \\
OVRL-V2 \cite{ovrlv2} & -- & 500 & 64.7 & 28.1 & -- & -- & -- & -- \\
OVRL & -- & 500 & -- & -- & -- & -- & 28.6 & 7.4 \\
\midrule
\multicolumn{9}{l}{\textit{Zero-shot but task-trained methods}} \\
ZSON \cite{zson} & CLIP & 500 & 25.5 & 12.6 & -- & -- & 15.3 & 4.8 \\
PSL \cite{psl} & CLIP & 500 & 42.4 & 19.2 & -- & -- & 18.9 & 6.4 \\
PixNav \cite{pixnav} & Foundation models & 500 & 37.9 & 20.5 & -- & -- & -- & -- \\
SGM \cite{sgm} & -- & 500 & 60.2 & 30.8 & -- & -- & 37.7 & 14.7 \\
VLFM \cite{vlfm} & BLIP-2 & 500 & 52.5 & 30.4 & 62.6 & 31.0 & 36.4 & 17.5 \\
\midrule
\multicolumn{9}{l}{\textit{Training-free and zero-shot methods}} \\
CoW \cite{cows} & CLIP & 500 & -- & -- & -- & -- & 9.2 & 4.9 \\
ESC \cite{esc} & GLIP-L + DeBERT & 500 & 39.2 & 22.3 & -- & -- & 28.7 & 14.2 \\
L3MVN \cite{l3mvn} & GPT-2 Large & 500 & 50.4 & 23.1 & 36.3 & 15.7 & -- & -- \\
VoroNav \cite{voronav} & GPT-3.5 & 500 & 42.0 & 26.0 & -- & -- & -- & -- \\
OpenFMNav \cite{openfmnav} & GPT-4/4V + G-SAM & 500 & 54.9 & 24.4 & -- & -- & -- & -- \\
\midrule
\multicolumn{9}{l}{\textit{Comparison with matched Qwen3-VL-Plus navigation components}} \\
WMNav$^{*}$ \cite{wmnav} & Qwen3-VL-Plus & 40 & 55.0 & 31.7 & 70.0 & 30.0 & 43.5 & 15.6 \\
WMNav + TM$^{*}$ & Qwen3-VL-Plus & 40 & 56.5 & 31.1 & 66.5 & 31.2 & 43.3 & 15.7 \\
\method{} & Qwen3-VL-Plus & 40 & \textbf{59.6} & \textbf{31.8} & \textbf{72.0} & \textbf{31.5} & \textbf{44.3} & \textbf{16.2} \\
\midrule
\multicolumn{9}{l}{\textit{Additional long-budget result}} \\
\method{} & Qwen3-VL-Plus & 500 & \textbf{65.3} & \textbf{32.1} & -- & -- & -- & -- \\
\bottomrule
\end{tabular}
\caption{ObjectNav results across HM3D and MP3D. Published rows provide cross-setting context; direct claims use the matched 40-step block. TM: Textual Memory; $^{*}$: our WMNav reproduction or extension; ``--'': unreported.}
\label{tab:objectnav_results}
\end{table*}

\section{Experiments}
\label{sec:experiments}

\subsection{Datasets and Evaluation Metrics}

\textbf{Benchmarks.}
We evaluate on the validation splits of HM3D v0.1, HM3D v0.2, and MP3D with scene-level sharding \cite{habitat_platform,chang2017matterport3d,ramakrishnan2021hm3d,batra2020objectnav}. HM3D v0.1 contains 2,000 episodes over 20 scenes and 6 goal categories; HM3D v0.2 contains 1,000 episodes with improved geometry and semantic annotations; MP3D contains 2,195 episodes over 11 scenes and 21 goal categories. \\
\textbf{Metrics.}
We report Success Rate (SR) and Success weighted by Path Length (SPL) \cite{anderson2018evaluation}:
\begin{equation}
\begin{aligned}
\mathrm{SR}
&=\frac{1}{N}\sum_{i=1}^{N}S_i,\\
\mathrm{SPL}
&=\frac{1}{N}\sum_{i=1}^{N}
S_i\frac{\ell_i^\star}{\max(\ell_i,\ell_i^\star)},
\end{aligned}
\end{equation}
where $N$ is the number of evaluated episodes, $S_i\in\{0,1\}$ is the success indicator for episode $i$, $\ell_i$ is its executed path length, and $\ell_i^\star$ is the shortest geodesic path to a valid goal. We additionally report paired backtracking, dead-end, and progress diagnostics where available.\\
\textbf{Cross-episode protocol.}
Episodes from each scene are assigned to one worker and executed consecutively. Scene memory starts empty, updates during navigation, and persists only within that scene; physical navigation state and short-term execution history are reset at every episode boundary. The memory-reset control uses identical workers, scene order, start poses, targets, action space, and step budget, but clears scene memory before every episode. The standard per-episode ObjectNav observations, actions, and success criterion are unchanged.\\
\subsection{Implementation Details}
Our controlled evaluations compare \method{} with WMNav$^{*}$, which resets scene memory and serves as the no-cross-episode-experience control. On HM3D, we additionally evaluate WMNav + Textual Memory$^{*}$, which retains VLM summaries of episode outcomes within each scene and contrasts free-form persistent summaries with the hierarchical visual-topological representation of \method{}.
Within each reported controlled comparison, the variants share the Qwen3-VL-Plus backbone \cite{qwen3vl}, candidate pipeline, action space, controller, and 40-step interaction budget. Following WMNav, the agent has radius 0.18~m and height 0.88~m, and receives $640\times480$ egocentric RGB-D observations with a $79^{\circ}$ field of view and $14^{\circ}$ downward camera pitch. We additionally report one 500-step HM3D v0.1 run as a long-budget diagnostic rather than a matched cross-episode comparison. \method{} maintains one scene memory per worker and uses fixed thresholds across datasets.
The Appendix provides prompts, lexical dictionaries, association and guard implementations, thresholds and update rules, tie-breaking, VLM settings, and pose-estimation details.

\subsection{Comparision with Others}
Table~\ref{tab:objectnav_results} includes published ObjectNav results for broader context, but differences in training regime, foundation model, interaction budget, and memory protocol preclude direct ranking against those rows. Among which WMNav + Textual Memory$^{*}$ retains free-form cross-episode summaries, and \method{} retains hierarchical visual-topological experience.
On HM3D v0.1, \method{} achieves 59.6 SR, 4.6 points above the memory-reset control and 3.1 points above persistent textual memory. On HM3D v0.2, it reaches 72.0 SR, corresponding gains of 2.0 and 5.5 points, with comparable SPL. These matched HM3D comparisons support room-owned visual-topological organization over both discarding experience and retaining it as free-form text. The separate 500-step HM3D v0.1 result of 65.3 SR and 32.1 SPL is a long-budget diagnostic and is not used for this controlled claim.
On MP3D, \method{} achieves 44.3 SR and 16.2 SPL, compared with 43.5/15.6 for the memory-reset WMNav$^{*}$ control and 43.3/15.7 for WMNav + Textual Memory$^{*}$. The improvement over both baselines extends the advantage of structured visual-topological experience beyond HM3D.


\begin{table}[t]
\centering
\small
\setlength{\tabcolsep}{3pt}
\begin{tabular}{cccccc}
\toprule
Topo & Persist. & Guard & Visual
& SR $\uparrow$ & SPL $\uparrow$ \\
\midrule
$\times$     & $\times$     & $\times$     & $\times$     & 55.0 & 31.7 \\
\midrule
$\times$     & $\checkmark$ & $\checkmark$ & $\checkmark$ & 56.5 & 31.1 \\
$\checkmark$ & $\times$     & $\checkmark$ & $\checkmark$ & 59.4 & 30.5 \\
$\checkmark$ & $\checkmark$ & $\times$     & $\checkmark$ & 58.0 & 30.0 \\
$\checkmark$ & $\checkmark$ & $\checkmark$ & $\times$     & 57.7 & 29.4 \\
\midrule
$\checkmark$ & $\checkmark$ & $\checkmark$ & $\checkmark$
& \textbf{59.6} & \textbf{31.8} \\
\bottomrule
\end{tabular}
\caption{Component ablation on HM3D v0.1. Topo.: room-topological structure; Persist.: scene-memory persistence; Guard: execution guard; Visual: visual object evidence, including in-room and remote-visible cues. Checkmarks indicate enabled components.}
\label{tab:ablation}
\end{table}

\subsection{Statistical Reliability and Cross-Scene Consistency}
On HM3D v0.2, \method{} achieves higher SR than WMNav + Textual Memory$^{*}$ in 21 of 36 scenes and ties in another 6, while obtaining higher SPL than WMNav$^{*}$ in 26 of 36 scenes. We further use a paired scene-cluster bootstrap with 10,000 replicates, resampling scenes while retaining all within-scene episodes and method pairings and recomputing episode-weighted metrics. The 95\% confidence intervals for \method{} minus WMNav$^{*}$ are $[-0.60,3.87]$ percentage points for SR and $[0.34,2.65]$ for SPL; against WMNav + Textual Memory$^{*}$, they are $[0.50,7.65]$ for SR and $[-1.82,2.24]$ for SPL. These results show cross-scene consistency and statistical support for the SR advantage over textual memory and the SPL advantage over the memory-reset control, while the remaining differences are interpreted as unresolved point-estimate trends. Additional scene-level results, bootstrap distributions, variance statistics, and implementation details of the statistical analysis are provided in the Appendix.

\begin{table}[t]
\centering
\small
\setlength{\tabcolsep}{3pt}
\begin{tabular}{lccc}
\toprule
Metric & \method{} & WMNav$^{*}$ & Change \\
\midrule
SR (\%) $\uparrow$             & 72.0 & 70.0 & +2.0 \\
Exclusive successes (\#)       & 112  & 92   & +20 \\
\midrule
Backtracks/episode $\downarrow$ & 1.49 & 1.53 & -2.6\% \\
Dead-end steps/episode $\downarrow$ & 3.11 & 3.21 & -3.3\% \\
Progress (m/step) $\uparrow$    & 0.80 & 0.78 & +1.7\% \\
\bottomrule
\end{tabular}
\caption{Paired HM3D v0.2 diagnostics. Outcomes use all 1,000 episodes; trajectories use the 608 joint successes. SR change is in percentage points, exclusive-success change is a count, and trajectory changes are relative to WMNav$^{*}$ and computed from unrounded values.}
\label{tab:paired_diagnostic}
\end{table}

\subsection{Ablation Study}

Table~\ref{tab:ablation} reports component-removal results under the 40-step HM3D v0.1 protocol. The full model obtains 59.6 SR and 31.8 SPL, compared with 55.0 and 31.7 for the memory-reset backbone. Removing topology reduces SR to 56.5. The no-persistence variant retains within-episode VTM but clears it at episode boundaries, yielding 59.4 SR and 30.5 SPL; this comparison isolates the contribution of carrying the hierarchy across episodes from its value as current-episode working memory. Removing the execution guard or visual object evidence yields 58.0/30.0 and 57.7/29.4 SR/SPL, respectively. These results isolate the guard at the component level, although lexical candidate re-ranking remains coupled with memory retrieval.

In the paired HM3D v0.2 diagnostics (Table~\ref{tab:paired_diagnostic}), \method{} obtains 72.0\% rather than 70.0\% SR and records 20 more exclusive successes. Among jointly successful episodes, it also has fewer backtracks and dead-end steps, with slightly higher progress per step. These trajectory differences are modest but consistent with a more directed search.


\subsection{Cross-Episode Progression}
To examine whether performance changes as cross-episode scene experience accumulates, we divide each scene's episode sequence into early and late halves and average the results across scenes. Table~\ref{tab:episode_progression} shows that \method{} gains 2.7 SR points from the early to late half while maintaining SPL. By comparison, the memory-reset WMNav$^{*}$ control and WMNav + Textual Memory$^{*}$ gain only 0.7 and 0.3 SR points, respectively, and neither improves SPL to the same extent. This pattern is consistent with hierarchical VTM harnessing a growing body of scene experience. However, because benchmark order, target composition, and start--goal difficulty are not controlled, the analysis is descriptive rather than causal.

\begin{table}[t]
\centering
\small
\setlength{\tabcolsep}{5pt}
\begin{tabular}{lcccccc}
\toprule
\multirow{2}{*}{Method} & \multicolumn{3}{c}{SR (\%) $\uparrow$} & \multicolumn{3}{c}{SPL (\%) $\uparrow$} \\
\cmidrule(lr){2-4}\cmidrule(lr){5-7}
& Early & Late & $\Delta$ & Early & Late & $\Delta$ \\
\midrule
WMNav$^{*}$ & 70.0 & 70.7 & +0.7 & 28.8 & 28.2 & -0.6 \\
WMNav + TM$^{*}$ & 67.7 & 68.0 & +0.3 & 31.2 & 31.1 & 0.0 \\
\method{} & 70.8 & 73.4 & +2.7 & 31.3 & 31.6 & +0.3 \\
\bottomrule
\end{tabular}
\caption{Cross-episode progression on HM3D v0.2. $\Delta$ is the mean per-scene late-minus-early change computed before rounding; episode order, target composition, and difficulty are not controlled. TM denotes Textual Memory.}
\label{tab:episode_progression}
\end{table}

\begin{figure}[htbp]
  \centering
  \textbf{HM3D v0.2, plant target}\\[0.3ex]

  \includegraphics[
    width=\linewidth,
    height=0.27\textheight,
    keepaspectratio
  ]{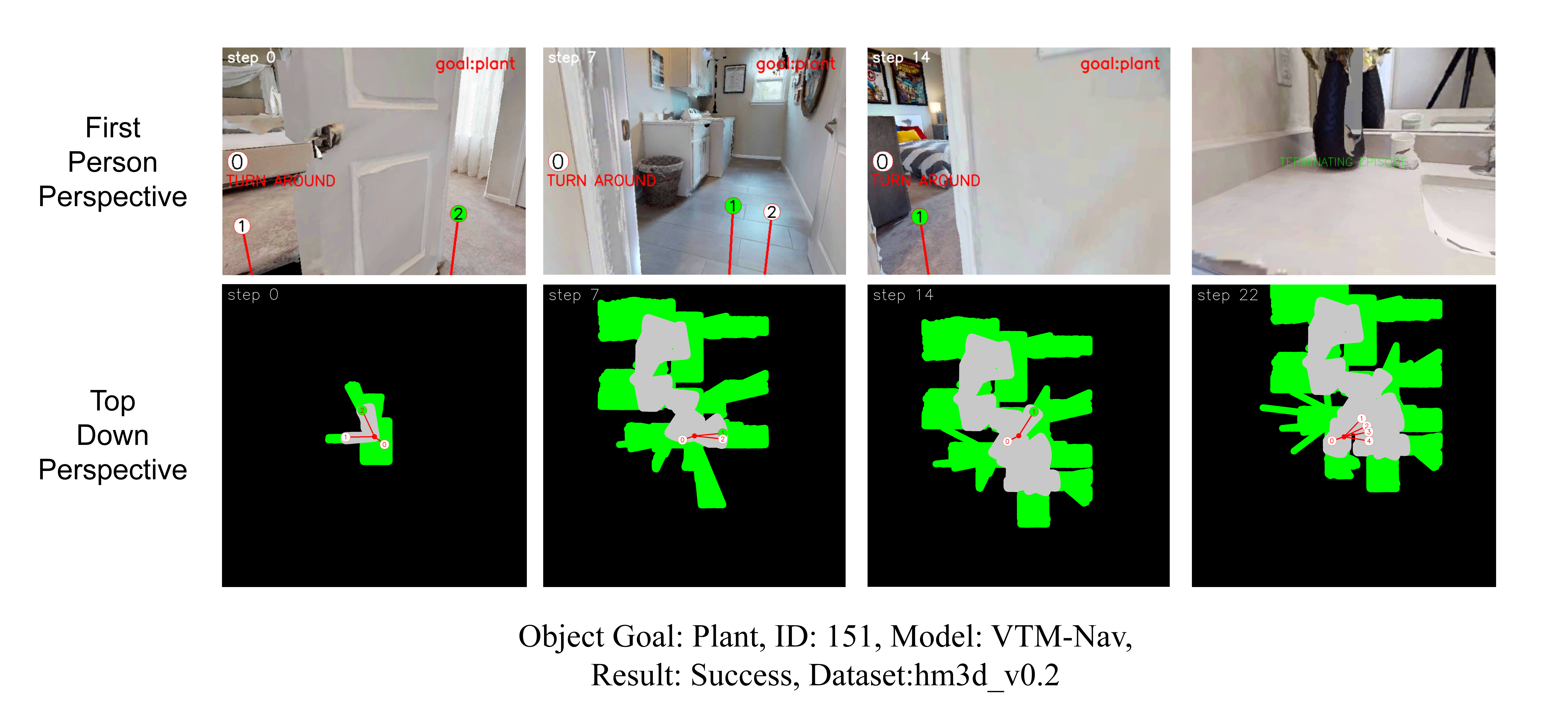}

  \par\vspace{0.6ex}

  \includegraphics[
    width=\linewidth,
    height=0.27\textheight,
    keepaspectratio
  ]{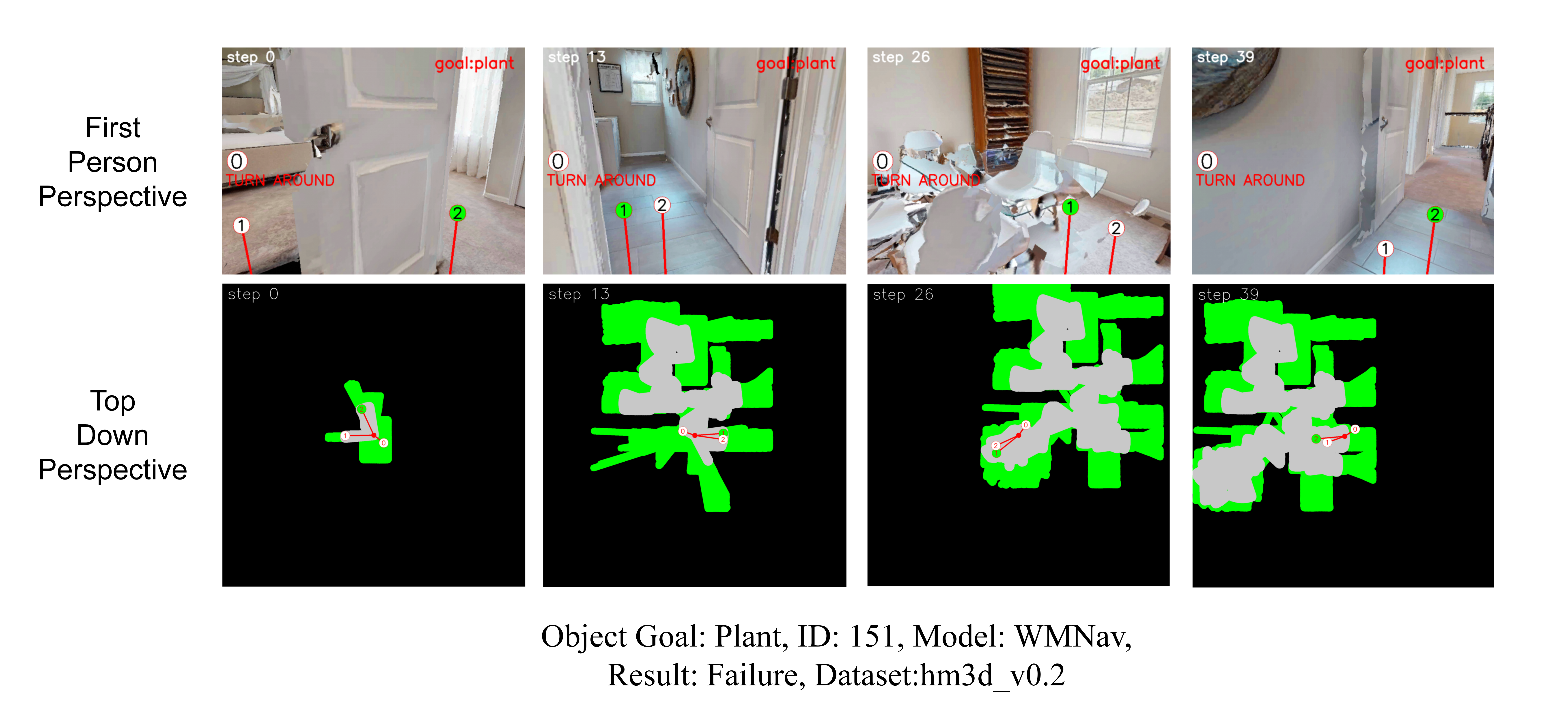}

  \caption{Failure-to-success comparison on HM3D v0.2. Top: \method{} reaches the target-bearing region and stops successfully. Bottom: WMNav continues searching until the interaction budget expires.}
  \label{fig:case_failure_success_comparison}
\end{figure}

\subsection{Qualitative Visualization and Case Study}

Figure~\ref{fig:case_failure_success_comparison} illustrates one failure-to-success pair for \method{} and WMNav. After local observations fail to reveal the target, WMNav continues exploring, whereas \method{} uses a retrieved room-level prior to reach the target-bearing region earlier and then stops from current visual evidence. This example illustrates a possible mechanism but is not a population-level error analysis.



\section{Conclusion}
\label{sec:conclusion}
We presented \method, a training-free framework for harnessing cross-episode experience across independently initialized ObjectNav requests. Its Hierarchical Visual-Topological Memory uses a coarse room graph to index fine-grained, room-owned visual experience, supporting re-localization and target-conditioned retrieval after episode reset while keeping the VLM and navigation components fixed. Under the controlled 40-step comparisons, \method{} improves SR point estimates over the matched memory-reset control on HM3D v0.1, HM3D v0.2, and MP3D, and over persistent textual memory on the two HM3D benchmarks, with comparable or higher SPL. These results support hierarchical visual-topological memory as a promising representation for cross-episode experience reuse without retraining or oracle scene knowledge.
\bibliographystyle{plainnat}

\bibliography{vtm_nav}

\end{document}